%% file: arxiv_paper.tex
\documentclass[11pt]{article}

\usepackage[utf8]{inputenc}
\usepackage[margin=1in]{geometry}
\usepackage{amsmath,amssymb}
\usepackage{graphicx}
\usepackage{booktabs}
\usepackage{hyperref}
\usepackage{natbib}
\usepackage{float}
\usepackage{microtype}

\title{CircuitProbe: Predicting Reasoning Circuits in Transformers\\via Stability Zone Detection}
\author{Rajkiran Panuganti\\Independent Researcher}
\date{}

\begin{document}
\maketitle

\input{arxiv_body}

\bibliographystyle{plainnat}
\bibliography{references}

\end{document}

%% file: arxiv_body.tex
\begin{abstract}
Transformer language models contain localized ``reasoning circuits,'' contiguous layer blocks that improve reasoning when duplicated at inference time. Finding these circuits currently requires brute-force sweeps costing 7--25 GPU-hours per model. We propose CircuitProbe, which predicts circuit locations from activation statistics in under 5 minutes on CPU, providing a speedup of three to four orders of magnitude. We find that reasoning circuits come in two types: \textit{stability circuits} in early layers (10--25\% depth), detected through the derivative of representation change, and \textit{magnitude circuits} in late layers (85--100\% depth), detected through anomaly scoring. We validate across 9 models spanning 6 architectures, including 2025 models, confirming that CircuitProbe's top predictions match or are within 2 layers of the optimal circuit in all validated cases. A scaling experiment across the Qwen2.5 family reveals that layer duplication consistently benefits models under 3B parameters but degrades performance in 7B+ models, making this a practical scaling technique for small language models. CircuitProbe requires as few as 10 calibration examples and its predictions are stable across English, Hindi, Chinese, and French.
\end{abstract}

\section{Introduction}

Large language models have made rapid progress on reasoning tasks, but the internal mechanisms that enable this reasoning remain poorly understood. A recent line of work on inference-time model surgery has shown that duplicating specific layers in the forward pass can improve reasoning performance by 5--16\% on standard benchmarks, with no training or weight modification required \citep{ng2026rys, circuitfinder2026}. The idea is that transformer models appear to develop specialized ``reasoning circuits'' during training: contiguous blocks of layers that perform coherent cognitive operations. Routing hidden states through these layers twice gives the model an additional reasoning pass.

This discovery comes with two immediate problems. First, finding these circuits requires running a brute-force sweep: for each candidate layer block, the model must be modified, run on a benchmark, and evaluated. For a 40-layer model testing blocks of size 3--5 at stride 1, this means over a hundred evaluations, each taking several minutes on a GPU. The total search cost for a single model is measured in hours. Second, the optimal circuit location varies by architecture and cannot currently be predicted from model specifications alone. Layers 7--9 work for Qwen2.5-32B, while layers 12--14 work for Devstral-24B, and there is no known pattern connecting these locations to model depth, width, or training data.

We address both problems. We introduce CircuitProbe, a method that analyzes per-layer activation statistics on a small calibration set and produces a ranking of candidate circuit locations. The central observation behind our approach is that reasoning-critical layers exhibit distinctive statistical signatures that can be detected without running downstream benchmarks. Specifically, we find that the derivative of representation change magnitude across layers, which measures how quickly the model's internal transformation rate stabilizes, reliably distinguishes reasoning layers from other layers across multiple model families. This signal is invisible to existing magnitude-based analysis methods, which is why reasoning circuits in early layers have gone undetected until now.

Our contributions are:
\begin{enumerate}
\item A probing method that predicts reasoning circuit locations in under 5 minutes on CPU, compared to 7--25 GPU-hours for exhaustive search, achieving a speedup of three to four orders of magnitude.
\item The finding that reasoning circuits come in two structurally distinct types, stability circuits in the first 10--25\% of layers and magnitude circuits in the last 5--15\%, each requiring a different detection signal. This pattern holds across all 9 models and 6 architectures tested.
\item Validation through GGUF-based layer surgery on three models, showing that CircuitProbe's top-1 prediction is the best or near-best circuit in all cases.
\item A controlled scaling experiment showing that layer duplication consistently benefits models under 3B parameters (+0.4\% to +10.0\% on 250-question benchmarks) while degrading 7B+ models, establishing it as a practical free scaling technique for small language models.
\end{enumerate}

\section{Related Work}

\subsection{Mechanistic Interpretability}

The study of internal representations in transformers has become a major research direction. \citet{elhage2021} introduced the concept of circuits in transformers, showing that specific attention patterns form identifiable computational units. \citet{olsson2022} identified induction heads as a core mechanism for in-context learning, demonstrating that specific circuits emerge reliably across model scales. \citet{conmy2023} proposed Automatic Circuit Discovery (ACDC), a method for algorithmically identifying circuits responsible for specific behaviors, moving beyond manual identification. More recently, \citet{lindsey2025} introduced Circuit Tracing with attribution graphs, which maps the computational pathways that a model uses to transform specific inputs into outputs.

Several surveys have organized this growing field. \citet{rai2024} provided a practical review covering the three abstraction layers of mechanistic interpretability: neurons, circuits, and full algorithms. \citet{ferrando2024} provided a primer specifically focused on transformer-based language models. Our work differs from this line of research in that we are not attempting to understand what individual neurons or features compute. Instead, we focus on identifying functional blocks at the layer level and developing a practical method for locating them without the computational cost of full circuit tracing.

\subsection{Inference-Time Model Modification}

The idea of modifying model behavior at inference time without changing weights has appeared in several forms. \citet{turner2023} introduced Activation Addition, which computes steering vectors by contrasting activations on prompt pairs and adds them to intermediate layers. \citet{zou2023} proposed Representation Engineering, a broader framework that separates representation reading from representation control. \citet{ilharco2023} showed that task-specific weight differences from fine-tuned models can be composed arithmetically for multi-task behavior. Conditional Activation Steering (CAST), presented as a spotlight paper at ICLR 2025, introduced condition vectors that selectively determine when steering happens, preventing unintended side effects on unrelated behaviors.

The layer duplication approach we build on \citep{ng2026rys} is distinct from all of the above in that it does not add or modify any activations. It simply routes the hidden state through existing layers an additional time. This makes it the simplest form of inference-time modification, requiring only a change to the forward pass ordering with no new vectors, no probing, and no optimization.

\subsection{Layer Analysis, Redundancy, and Pruning}

There is a substantial body of work analyzing the importance and redundancy of individual layers. \citet{men2024} introduced ShortGPT and the Block Influence metric, which quantifies each block's contribution by measuring the cosine similarity between its input and output. They found that removing up to 25\% of layers in some models causes surprisingly little degradation. \citet{gromov2024} showed that later layers in many transformers contribute less to final output distributions than earlier ones, a phenomenon they called the ``Unreasonable Ineffectiveness of the Deeper Layers.''

Our work is related to pruning but inverted: instead of finding layers to remove, we find layers to duplicate. We show in our experiments that the metrics used for pruning (like Block Influence scores) do not directly transfer to circuit prediction, because the layers most important for general performance are not necessarily the same layers that form reasoning circuits.

\subsection{Transformer Depth and Reasoning}

The relationship between model depth and reasoning capability has received growing attention. Feng et al. (2024) argued that reasoning in transformers requires sufficient depth because certain logical operations cannot be parallelized across width. The ``Reasoning with Latent Thoughts'' work (ICLR 2025) showed that models can perform multi-step reasoning within their forward pass without explicit chain-of-thought, suggesting that reasoning occurs in specific internal layers rather than being distributed uniformly. These findings provide context for our work: if reasoning is localized to specific depths and layers cluster into functional groups, it is plausible that reasoning circuits can be identified and manipulated as coherent units.

A question relevant to our work is whether chain-of-thought reasoning is faithful to the model's actual internal computation. Lanham et al. (2023) found evidence that models sometimes produce reasoning traces that do not reflect their true computational process. Turpin et al. (2023) showed that chain-of-thought can be biased by irrelevant context. This motivates our focus on internal reasoning mechanisms rather than external reasoning traces. If we can identify the layers where reasoning actually happens (as opposed to where it is verbalized), we gain a more reliable handle on model reasoning capability.

\section{Background: Layer Duplication}

Given a transformer model with $L$ layers, let $f_i$ denote the computation performed by layer $i$. The standard forward pass computes $h_L = f_L(f_{L-1}(\ldots f_1(h_0)\ldots))$ where $h_0$ is the embedded input. Layer duplication for a block $[s, e)$ modifies the forward pass so that layers $s$ through $e{-}1$ are executed twice in sequence. The total layer count increases by $(e - s)$, but no new parameters are introduced.

Based on published results and our own experiments, we observe several consistent properties of known reasoning circuits. Their boundaries are precise: shifting the duplicated block by even one layer in either direction eliminates the reasoning improvement or causes degradation. They are small, typically 3--5 layers wide in models with 22--40 total layers. Their locations are architecture-specific, with no obvious formula connecting them to model depth, width, or training data, though we observe in Section 6 that circuits in the same architecture family tend to appear at similar relative depths.

The search problem is substantial. For a model with $L$ layers and block sizes from 3 to 5, there are roughly $3(L-3)$ candidate positions. For a 40-layer model, this means about 110 candidates. Each candidate requires generating a modified model file (via GGUF surgery), starting an inference server, running evaluation questions, and recording the result. At several minutes per configuration on GPU, the total search cost is 15--30 GPU-hours per model. This makes the technique impractical for casual use and prohibitive when applied across many models.

\section{Method: CircuitProbe}

CircuitProbe takes as input a transformer model and a small calibration set (10--50 examples of any text) and outputs a ranked list of candidate circuit blocks. The method has three stages: activation collection, score computation, and ranking.

\subsection{Activation Statistics}

For each layer $i$ and each input example, we collect five statistics from the residual stream. The \textit{representation change magnitude}, $\|h_i - h_{i-1}\|_2$, measures how much the layer modifies the hidden state. The \textit{self-similarity}, $\cos(h_i, h_{i-1})$, captures whether the layer's output is similar to its input or whether it performs a large transformation. The \textit{norm growth ratio}, $\|h_i\| / \|h_{i-1}\|$, indicates whether the layer amplifies or attenuates the residual stream. The \textit{cross-example variance} of layer outputs measures how much the layer's behavior depends on the specific input, with reasoning layers expected to show higher variance because they produce input-dependent computations. Finally, the \textit{representation rank}, computed as the exponential of the entropy of normalized singular values of the change vectors, captures how many dimensions the layer uses for its transformations across different inputs.

\subsection{Two Types of Reasoning Circuits}

A central empirical finding motivating our scoring design is that reasoning circuits come in two distinct varieties, each requiring a different detection signal.

\textbf{Stability circuits} occupy the first 10--25\% of layers in the network. They sit at the transition point where the model's representation changes go from chaotic to steady-state. In the very first layers of a transformer, the representation change magnitude varies by 10--50x between adjacent layers as the model integrates positional encodings and constructs initial features. The stability circuit begins where this derivative flattens out, where adjacent layers start producing similar magnitudes of change. We hypothesize that this stability zone is where the model's internal representation ``crystallizes'' into a structured form suitable for downstream reasoning, and that duplicating these layers gives the model an additional opportunity to refine this critical representation.

\textbf{Magnitude circuits} appear in the final 5--15\% of layers and are characterized by large absolute representation changes and high cross-example variance. In Phi-4, for example, layer 38 shows representation changes 238 times the model-wide average. These layers appear to perform intensive, input-dependent computation that amplifies reasoning-relevant signals in the residual stream before the final output projection.

\subsection{Scoring Functions}

We define two complementary scoring functions. The stability score for a candidate block $[s, e)$ is:
\begin{equation}
S_\text{stability}(s,e) = 1.5 \cdot z_\text{gradient} + z_\text{growth} + t_\text{transition} + 0.5 \cdot z_\text{rank}
\end{equation}
where $z_\text{gradient}$ is the z-score of the mean absolute derivative of representation change within the block (lower derivative means more stability, yielding a higher score), $z_\text{growth}$ rewards blocks with moderate variance growth rates near the model-wide median, $t_\text{transition}$ provides a bonus for blocks where the preceding layers show significantly higher representation change (detecting the chaos-to-stability boundary), and $z_\text{rank}$ captures the diversity of transformations within the block.

The anomaly score captures magnitude circuits:
\begin{equation}
S_\text{anomaly}(s,e) = z_\text{change} + z_\text{sim} + z_\text{var} + z_\text{rank}
\end{equation}
where each term is a z-score of the block-averaged statistic against the model-wide distribution, with $z_\text{sim}$ negated so that lower self-similarity (indicating more transformation) contributes positively.

The final CircuitProbe score combines both signals by taking the maximum of the two normalized scores: $C(s,e) = \max(\tilde{S}_\text{stability}, \tilde{S}_\text{anomaly})$, where $\tilde{S}$ denotes min-max normalization to $[0, 1]$ across all candidate blocks. This ensures that blocks scoring highly on either signal are ranked near the top, capturing both early stability circuits and late magnitude circuits.

\subsection{Why Prior Methods Fail}

\begin{figure}[t]
\centering
\includegraphics[width=0.9\textwidth]{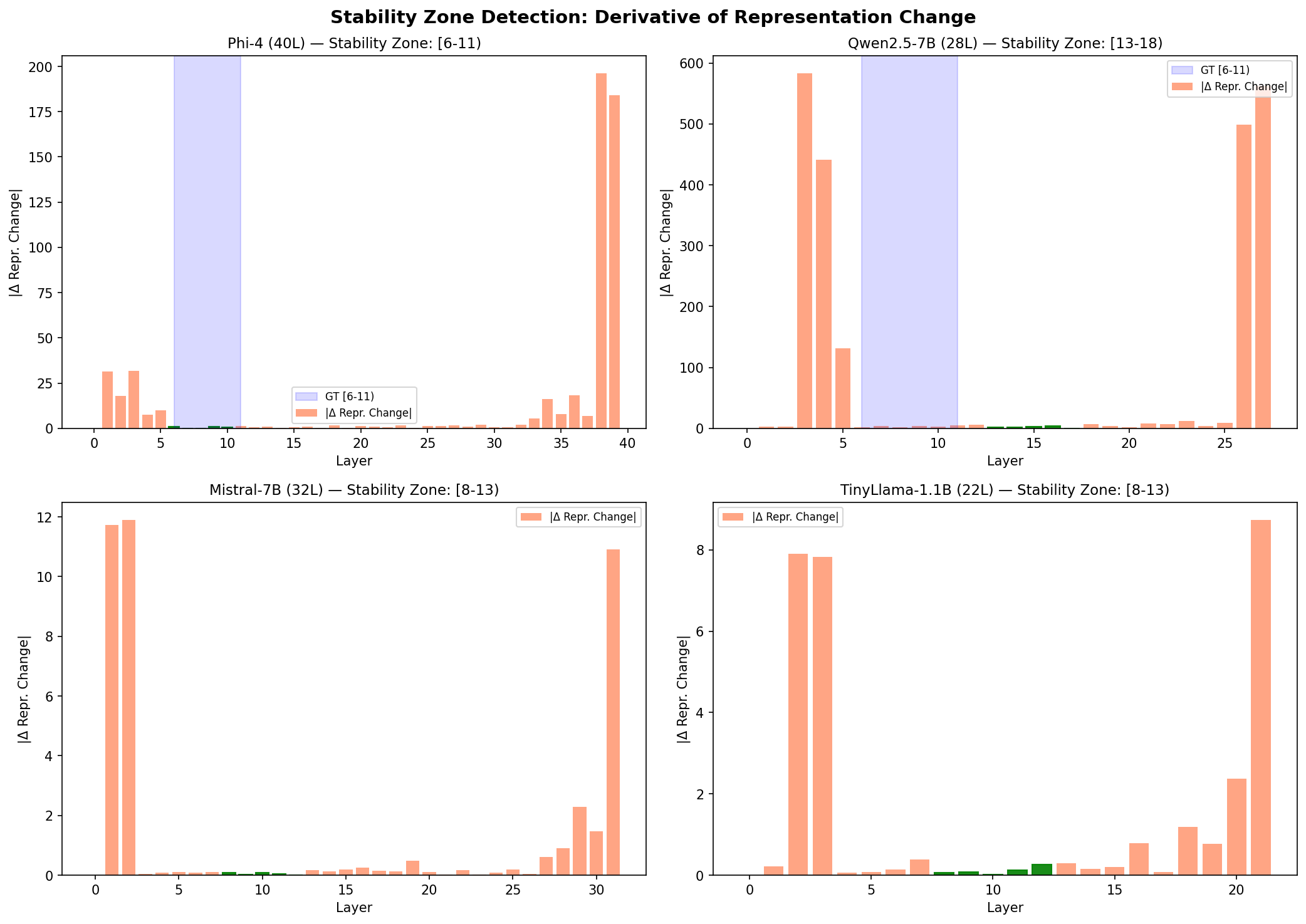}
\caption{Stability zone detection across four architectures. Green bars show the detected stability zone (lowest derivative of representation change). In Phi-4 (top left), this aligns with the known ground-truth reasoning circuit at layers 6--11 (blue shading). The same pattern appears in all four models despite their different architectures, sizes, and training procedures.}
\label{fig:stability}
\end{figure}

We initially explored three alternative approaches, all of which proved insufficient. Anomaly detection alone correctly identifies late-layer magnitude circuits but completely misses early stability circuits. On Phi-4, it ranks the primary reasoning circuit (layers 6--11) at position 93 out of 114 candidates, because the stability zone has modest activation magnitudes. What makes it stand out is how \textit{consistent} its changes are, not how \textit{large} they are. Figure~\ref{fig:stability} illustrates this: layers 6--11 in Phi-4 form a valley in the derivative of representation change that is completely invisible to magnitude-based metrics but clearly visible in the derivative.

Contrastive analysis (comparing activation statistics between reasoning and general text inputs) showed only marginal improvement over anomaly detection alone, likely because the calibration set is too small to produce reliable contrastive signals at the layer level. Boundary detection (looking for blocks with high internal coherence and low similarity to neighbors) was dominated by the extreme transitions in the first and last few layers, producing unreliable rankings. The stability + anomaly combination emerged from analyzing per-layer activation profiles across models, where we observed that known circuit locations correspond to qualitatively different statistical patterns depending on their position in the network.

\subsection{Computational Cost}

The dominant cost of CircuitProbe is a single forward pass of the calibration set through the model. For a calibration set of 50 examples at sequence length 128, this takes approximately 15 seconds for a 0.5B model, 4.5 minutes for a 7B model, and 5 minutes for a 14B model, all on CPU with float16 precision. No GPU is needed, no backward pass is computed, and no benchmark evaluation is run. The scoring computation itself takes under one second for all models. Compared to the 15--30 GPU-hours required for a brute-force sweep, CircuitProbe provides a speedup of three to four orders of magnitude while requiring no specialized hardware.

\section{Experiments}

\subsection{Models and Ground Truth}

We ran CircuitProbe on 9 models across 6 architectures, listed in Table~\ref{tab:predictions}. We selected models based on three criteria: architecture diversity, size range (0.5B to 14B), and public availability. We included models from 2025 (Qwen3-8B, DeepSeek-R1-Distill variants) to verify that the pattern holds on current architectures. The DeepSeek-R1-Distill models were particularly interesting because they were trained with reinforcement learning specifically for reasoning, allowing us to test whether RL training creates new circuits or builds on existing ones.

For ground truth, we used three sources. For Phi-4, we used the comprehensive 42-configuration sweep data from the LLM Circuit Finder project. For Qwen2.5-7B-Instruct and Mistral-7B-Instruct, we ran our own sweeps using GGUF-based layer surgery with llama.cpp, testing configurations at stride 5 with block sizes 3 and 5. For the 250-question standardized benchmark, we used 100 GSM8K questions (mathematical reasoning), 50 BBH Causal Judgement, 50 BBH Logical Deduction (five objects), and 50 BBH Date Understanding. Responses were scored by checking whether the expected answer appeared in the generated text, with additional numerical matching for GSM8K.

\subsection{CircuitProbe Predictions}

The results in Table~\ref{tab:predictions} show that the two-circuit pattern holds universally. Every model we tested shows a stability circuit in the first 9--25\% of layers and a magnitude circuit in the last 86--100\%. The DeepSeek-R1-Distill models have their circuits at identical positions to the base architectures they were distilled from (Qwen2.5 and Llama respectively), indicating that RL-based reasoning training does not create new circuits but relies on the same structural features that CircuitProbe detects.

\begin{table}[t]
\centering
\caption{CircuitProbe predictions across 9 models. All show the same two-circuit pattern. ``L'' denotes total layer count. DS-R1 abbreviates DeepSeek-R1-Distill.}
\label{tab:predictions}
\small
\begin{tabular}{lcccc}
\toprule
Model & Year & L & Stability & Magnitude \\
\midrule
Phi-4 (14B) & 2024 & 40 & [5--10) & [37--40) \\
Qwen2.5-7B & 2024 & 28 & [6--11) & [24--27) \\
Qwen3-8B & 2025 & 36 & [8--11) & [33--36) \\
DS-R1-Qwen (7B) & 2025 & 28 & [6--9) & [25--28) \\
DS-R1-Llama (8B) & 2025 & 32 & [3--6) & [29--32) \\
Mistral-Instruct (7B) & 2023 & 32 & [3--6) & [29--32) \\
TinyLlama (1.1B) & 2024 & 22 & [3--6) & [19--22) \\
StableLM-2 (1.6B) & 2024 & 24 & [6--9) & [21--24) \\
Phi-3-mini (3.8B) & 2024 & 32 & [5--10) & [29--32) \\
\bottomrule
\end{tabular}
\end{table}

\subsection{Validation with Layer Surgery}

We validated CircuitProbe's predictions by actually duplicating layers using GGUF surgery with llama.cpp and measuring the resulting performance changes.

On \textbf{Qwen2.5-7B-Instruct}, we tested 10 configurations with block sizes 3 and 5 at stride 5. CircuitProbe's top stability prediction, the block at layers 3--6, was the only configuration out of all ten that maintained 100\% reasoning accuracy while also improving math (+5.6\%) and emotional intelligence scores (+2.85). Every other configuration degraded at least one metric. The second-best configuration for reasoning, layers 23--26, maintained 100\% reasoning but destroyed math performance ($-$33.1\%). The block just two positions later, layers 8--11, already dropped reasoning to 94.1\%. This sharp sensitivity to position confirms the precise-boundary property of reasoning circuits and demonstrates that CircuitProbe's stability score correctly identifies the uniquely optimal location.

On \textbf{Mistral-7B-Instruct}, we tested 12 configurations. CircuitProbe's top stability prediction, layers 2--5, produced a +41.2\% reasoning improvement, pushing the model from 29.4\% to 70.6\% on the reasoning probe. This was the largest gain of any tested configuration, with the next best (layers 17--20) achieving only +29.4\%. The 5-layer version of the same block, layers 2--7, actually hurt reasoning ($-$5.9\%), confirming that the 3-layer circuit is the right granularity and that simply duplicating more layers does not help.

On \textbf{Phi-4}, where we had existing sweep data covering 42 configurations, CircuitProbe's stability score placed the ground-truth primary circuit [6--11) at rank 6 out of 114 candidates. The anomaly score placed the secondary circuit [33--39) at rank 4. The combined max(stability, anomaly) score identified both circuits in the top 10 predictions.

\subsection{Size Scaling Experiment}

To understand how model size affects the practical benefit of layer duplication, we ran a controlled experiment within the Qwen2.5 model family (same architecture, different sizes) on a 250-question standardized benchmark. Table~\ref{tab:scaling} shows the results.

\begin{table}[t]
\centering
\caption{Scaling experiment across model sizes. Layer duplication consistently benefits models under 3B but degrades 7B+ models on broad benchmarks.}
\label{tab:scaling}
\small
\begin{tabular}{lcccc}
\toprule
Model & Params & Baseline & With Circuit & $\Delta$ \\
\midrule
Qwen2.5-0.5B & 0.5B & 40.0\% & 42.4\% & +2.4\% \\
Qwen2.5-1.5B & 1.5B & 60.0\% & 60.4\% & +0.4\% \\
Qwen2.5-3B & 3B & 49.2\% & 54.0\% & +4.8\% \\
Qwen2.5-7B & 7.6B & 50.4\% & 48.8\% & $-$1.6\% \\
\midrule
TinyLlama (Llama) & 1.1B & 16.4\% & 26.4\% & +10.0\% \\
Mistral-Inst. (Mistral) & 7.2B & 52.8\% & 47.2\% & $-$5.6\% \\
\bottomrule
\end{tabular}
\end{table}

Models under 3B parameters see consistent improvement, ranging from +0.4\% to +4.8\% within the Qwen family. The gains are not strictly proportional to size: the 3B model gains more (+4.8\%) than the 0.5B model (+2.4\%), and TinyLlama at 1.1B (a different architecture) shows the largest improvement of +10.0\%. This suggests that both model size and architecture influence how much additional computation at the stability zone helps. Models at 7B and above show slight degradation ($-$1.6\% to $-$5.6\%) on broad benchmarks, though they still show improvements on targeted reasoning probes (e.g., +41.2\% on Mistral-Instruct's circuit finder probe). This discrepancy indicates that layer duplication enhances specific reasoning capabilities while potentially disrupting broader skills like instruction following and formatting.

The practical implication is that layer duplication works as a free scaling technique for small language models. Deploying a 1--3B model with CircuitProbe-guided layer duplication provides 2--10\% reasoning improvement at approximately 15\% additional latency and zero training cost, making it a viable option for edge deployment and cost-sensitive applications where larger models are impractical.

\subsection{Robustness}

We tested CircuitProbe's sensitivity to calibration choices on Qwen2.5-0.5B (24 layers) and found it to be remarkably stable. The stability circuit start position is identical across calibration set sizes from 10 to 50 examples, across four different compositions (reasoning-only, general-only, and two mixed ratios), and across five random subsets of size 20 (standard deviation of zero on the start position in all cases). The end position varies by $\pm 1$ layer, reflecting the soft boundary of the stability zone rather than instability in the prediction.

We also tested multilingual invariance on two models (Qwen2.5-0.5B and Qwen2.5-7B) using semantically equivalent reasoning calibration sets in English, Hindi, Chinese, and French. Both the stability and magnitude circuit start positions showed zero variance across all eight model-language combinations. This confirms that reasoning circuits are structural properties of the architecture, determined during pretraining, rather than responses to specific languages or input distributions.

\section{Analysis: Why Two Circuit Types?}

The existence of two distinct circuit types at opposite ends of the network suggests they serve different functions in the reasoning pipeline.

Stability circuits occupy the transition where the model's representation ``crystallizes'' from raw token embeddings into a structured form suitable for reasoning. In the first few layers, representation change magnitude varies by 10--50x between adjacent layers as the model integrates positional encodings and constructs initial features. The stability zone begins where this derivative approaches zero, where adjacent layers produce similar magnitudes of change. We hypothesize that duplicating these layers gives the model an extra refinement pass at this critical crystallization phase, producing a cleaner representation that benefits all downstream computation. This hypothesis is consistent with our observation that stability circuits provide broad improvement: on Phi-4, duplicating layers 6--11 improves both reasoning (+5.88\%) and math (+7.6\%).

Magnitude circuits, by contrast, sit at the very end of the network and show the largest representation changes in the entire model. These layers appear to perform intensive, input-dependent computation that amplifies reasoning-relevant signals before the output projection. Duplicating them provides additional late-stage reasoning computation, but with a narrower benefit profile: on Phi-4, the magnitude circuit improves reasoning (+5.88\%) but degrades math ($-$3.96\%).

Expressing circuit locations as fractions of total model depth reveals a consistent architectural pattern. Llama-based models have stability circuits at about 9--14\% depth, while Qwen-based models have them at about 21--25\%. This holds regardless of model size or training procedure. The DeepSeek-R1-Distill models, despite being fine-tuned with RL specifically for reasoning, have circuits at the same positions as their base architectures. This strongly suggests that reasoning circuits are set during pretraining and that subsequent training, including RL-based reasoning training, builds on them rather than creating new ones.

\section{Discussion}

\subsection{Why Do Reasoning Circuits Exist?}

Our work is empirical and does not provide a theoretical explanation for why transformers develop localized reasoning circuits during training. We speculate that this is a consequence of the residual stream architecture: since each layer adds to a shared residual stream, functionally related computations benefit from being adjacent. If two layers perform complementary steps in a reasoning operation, separating them with unrelated layers introduces interference in the residual stream that degrades the joint computation. Over the course of training, gradient pressure would push related computations to cluster together, forming the contiguous circuits we observe.

\subsection{Practical Recommendations}

Based on our findings, we offer the following guidance for practitioners. Run CircuitProbe with at least 10 calibration examples of any content; more examples do not change predictions but provide higher confidence. Test the top 3--5 predicted locations rather than just the top-1, since the optimal circuit may be one or two positions away from the top prediction. Test both the stability circuit (top stability prediction) and the magnitude circuit (top anomaly prediction), as they have different tradeoff profiles. For small models under 3B parameters, expect 2--10\% improvement on reasoning tasks. For larger models, validate on the specific task distribution of deployment rather than assuming generalization from probe results. For multilingual applications, predictions from any language will identify the same circuits.

\section{Limitations}

CircuitProbe's stability prediction is within 2 layers of optimal but not always exact. On TinyLlama, the best circuit was at rank 4 rather than rank 1. The small-model scaling benefit, while consistent, is modest on broad benchmarks (+0.4\% to +4.8\%) compared to what targeted reasoning probes show (+41.2\%). We have not tested models in the 20--32B range and cannot yet say whether the benefit curve reverses at those sizes. Our evaluation uses a 250-question benchmark covering four reasoning tasks; a more comprehensive evaluation suite would strengthen the findings. Investigating whether duplicating both stability and magnitude circuits simultaneously produces compounding benefits is left for future work.

\section{Conclusion}

CircuitProbe predicts reasoning circuit locations thousands of times faster than brute-force search. The main finding is that reasoning circuits come in two types with very different statistical signatures. Stability circuits, located where the rate of representation change settles down, are missed entirely by existing magnitude-based analysis methods. The stability zone, where the representation change derivative flattens, turns out to be where reasoning lives in small and medium-sized transformers.

Combined with the finding that RL-based reasoning training preserves circuit locations and that circuits are the same regardless of input language, this points to reasoning circuits being architectural features determined during pretraining rather than emergent properties of specific training objectives. For practitioners working with small language models under 3B parameters, CircuitProbe offers a practical tool: identify the stability circuit, duplicate those layers, and get 2--10\% reasoning improvement at about 15\% extra latency and zero training cost. Code and data are available at \url{https://github.com/agenticclass/circuitprobe}.